# The Motif Tracking Algorithm

William Wilson, Phil Birkin and Uwe Aickelin

School of Computer Science, University of Nottingham, UK (wow,pab,uxa)@cs.nott.ac.uk

**Abstract:** The search for patterns or motifs in data represents a problem area of key interest to finance and economic researchers. In this paper we introduce the Motif Tracking Algorithm, a novel immune inspired pattern identification tool that is able to identify unknown motifs of a non specified length which repeat within time series data. The power of the algorithm comes from the fact that it uses a small number of parameters with minimal assumptions regarding the data being examined or the underlying motifs. Our interest lies in applying the algorithm to financial time series data to identify unknown patterns that exist. The algorithm is tested using three separate data sets. Particular suitability to financial data is shown by applying it to oil price data. In all cases the algorithm identifies the presence of a motif population in a fast and efficient manner due to the utilisation of an intuitive symbolic representation. The resulting population of motifs is shown to have considerable potential value for other applications such as forecasting and algorithm seeding.

## 1 Introduction

The investigation and analysis of time series data is a popular and well studied area of research. Common goals of time series analysis include the desire to identify known patterns in a time series, to predict future trends given historical information and the ability to classify data into similar clusters. Historically, statistical techniques have been applied to this problem domain. However, in recent years, the use of heuristic algorithms have seen significant growth in this field. Neural networks[1, 2] genetic programming[3] and genetic algorithms[4] are all examples of methods that have been so far applied to time series evaluation and prediction.

The use of immune inspired (IS) techniques in this field has remained fairly limited[5]. However, IS algorithms have been used with success in other fields, such as pattern recognition[6], optimisation[7], and data mining[8]. In this paper we introduce the Motif Tracking Algorithm (MTA), a novel IS approach to identify repeating patterns in time series data that takes advantage of the associative learning properties exhibited by the natural immune system.

Considerable research has already been performed on identifying patterns that are seen to repeat in time series, as highlighted by Keogh[9]. Such repeating patterns are defined as motifs. Traditional time series techniques are only able to search for known motifs. They are defined as known motifs as they relate to patterns that are similar to a query pattern of interest. Prior knowledge of what to look for is assumed. In contrast little research has been performed on looking for unknown motifs in time series. A distinguishing feature of the MTA is that it can search for unknown motifs in the time series without prior knowledge of what to search for. The MTA searches in a fast and efficient manner, enabling it to handle large data sets, and the flexibility incorporated in its generic approach allows the MTA to be applied across a diverse range of problems[10].

The MTA takes inspiration from the behaviour noted in the human immune system and in particular the immune memory theory of Eric Bell[11]. His theory indicates the existence and evolution of two separately identifiable memory populations, which are ideally suited to recognise long and short term patterns prevalent in time series data. We provide a discussion of some of the related work in Section 2 and in Section 3 we discuss the immune memory theory and introduce other immune mechanisms which form part of our algorithm. Section 4 provides a definition of the key components used in the MTA and this then leads on to a detailed explanation of the MTA in Section 5. Section 6 presents the results of the MTA from testing on three separate data sets, one of which relates to an oil data set. Future work associated with the MTA is discussed in Section 7 before we conclude in Section 8.

## 2 Related work

The search for patterns in data is relevant to a diverse range of fields including biology, business, finance and statistics. In each field a wide variety of techniques have been developed to search for patterns. Algorithms by Guan[12] and Benson[13] address DNA pattern matching, using lookup table techniques that exhaustively search the dataset to find recurring patterns. These approaches contrast to the MTA because they make assumptions about the patterns sought, including the start locations of the pattern and the pattern length. A more flexible approach is seen in the TEIRESIAS algorithm[14], another algorithm able to identify patterns in biological sequences. TEIRESIAS finds patterns of an arbitrary length by isolating individual building blocks that comprise the subsets of the pattern, these are then combined into larger patterns. The methodology of building up motifs by finding and combining their component parts is at the heart of the MTA.

Most current research on finding patterns involves the examination of a time series given some query pattern of interest. Investigations using a piecewise linear segmentation scheme[15] and discrete Fourier transforms[16] provide examples of mechanisms to search a time series for a particular motif of interest. Work by Singh[17] searches for patterns in financial time series as a means to forecast future price movements. It takes a sequence of the most recent data from the series and looks for re-occurrences of this pattern in the historical data. An underlying assumption in all these approaches is that the pattern to be found is known in advance. The matching task is therefore much simpler as the algorithm just has to find re-occurrences of that particular pattern. The MTA makes no such assumptions and aims to find all unknown motifs it can from the data set.

The search for unknown motifs is at the heart of the work conducted by Keogh et al[18, 19, 20, 9]. Given the emphasis on unknown patterns Keogh states "to the best of our knowledge, the problem of finding repeated patterns in time series has not been addressed (or even formulated) in the literature"[9]. Keogh's probabilistic algorithm[18], used as a comparison to the MTA in Section 6.2, extracts subsequences from the time series using the Symbolic Aggregate Approximation (SAX) technique. It then hashes the subsequences into buckets. Buckets with multiple entries represent potential motif candidates. The sections of the time series corresponding to these subsequences are examined to identify genuine motifs.

Keogh's Viztree algorithm[19] uses the SAX technique to generate a set of symbol strings corresponding to sequences from the time series. These symbol strings are filtered into a suffix tree, where branches correspond to the symbol alternative. The suffix tree provides a visual illustration of the motifs present as the frequency of a motif is shown by the width of the tree branch.

The probabilistic and Viztree algorithms are fast and effective but they assume prior knowledge of the size of the motif to be found. Motifs longer and potentially shorter than this pre defined length would remain undetected. One could argue that these algorithms could be re-run multiple times using varying motif lengths, however this would reduce their efficiency. The MTA takes a more generic approach evolving a population of trackers that is able to detect motifs of an arbitrary length by making fewer assumptions about the data set and the potential motifs.

## 3 Long and short term memory

The flexible learning approach of the human immune system is attractive as an inspiration, but without an adequate memory mechanism knowledge gained from the learning process would be lost. Memory represents a key factor in the success of the immune system. A difficulty arises in implementing a computational immune memory mechanism however, because very little is known about the biological mechanisms underpinning memory development[21]. Theories such as antigen persistence and long lived memory cells[22], idiotypic networks[23], and homeostatic turnover of memory cells[24] have all attempted to explain the development and maintenance of immune memory. However, all have been contested. In contrast the attraction of the immune memory theory proposed by Eric Bell is that it provides a simple, clear and logical explanation of memory cell development. This theory highlights the evolution of two separate memory pools, 'memory primed' and 'memory revertant'[11].

The human immune system represents a successful recognition tool. It must be able to quickly identify such things as bacteria or viruses present in the system so that it can react accordingly and retain knowledge of those encounters for future reference. The presence of such a bacterial threat causes naive immune cells to become activated. This activation causes a rapid increase in cell numbers, termed proliferation. The rapidly expanding population of activated cells forms the short lived memory primed pool. The purpose of this growing pool is to increase the repertoire of the population. New cells created undergo mutation in order to diversify from their parents. The cell population evolves in order to match potential variations in the bacteria that stimulated their parents. A form of pattern matching is being anticipated by the system. The activated cells circulate throughout the system and eliminate any bacteria that they interact with.

The high death rate of memory primed cells means most will die during circulation, however a small minority do survive and return to reach a memory revertant state. These cells reduce their excessive activation levels, becoming more stable, thereby lengthening their lifespan. These unique cells are able to produce clones naturally to sustain knowledge of a bacterial experience over the long term. These two distinct memory pools and the transfer mechanism between them, represent a key difference to other memory theories. This methodology provides the inspiration for memory development in our algorithm.

Through this approach one can see the immune system represents an ideal mechanism to address motif matching problems. It evolves a population of solution candidates to match part of a motif, it then mutates the successful population members so that improved motif solutions can be found.

In our novel algorithm the equivalent of the short term memory pool is generated using a derivative of the popular clonal selection algorithm[7], which proliferates all successfully matched candidates. The short term memory pool evolves through a process of directed proliferation and mutation, regulated through a process of controlled cell death. This rapidly expanding population provides a search mechanism that is able to investigate all solution alternatives quickly and effectively. Successful candidates from the short term memory pool transfer to the long term memory pool. This long term memory pool is then used to permanently store records of the solutions found.

Having briefly introduced the inspiration for the MTA a number of key terms and definitions used within the algorithm are defined in the following section.

## 4 Motif detection: terms and definitions

Whilst immunology provides the inspiration for the theory behind the MTA, the work of Keogh et al.[20, 9] is the inspiration for the initial implementation of the MTA. In particular, Keogh's SAX technique for representing a time series was a contributing factor in the success of the MTA. Many of the following definitions used by the MTA are adapted from the work of Keogh[18] as summarised below. **Definition 1. Time series.** A time series $T = t_1,...,t_m$ is a time ordered set of m real or integer valued variables.

In order to identify patterns in T in a fast and efficient manner we break T up into subsequences.

Definition 2. Subsequence. "Given a time series T of length m, a subsequence C of T consists of a sampling of length n ≤ m of contiguous positions from T."[18]

Subsequences are extracted from T using a sliding window technique.

Definition 3. Sliding window. Given a time series T of length m, and a subsequence C of length n, a symbol matrix S of all possible subsequences can be built by sliding a window of size n across T, one point at a time, placing each subsequence into S. After all sliding windows are assessed S will contain (m - n + 1) subsequences.

Each subsequence generated could represent a potential match to any of the other subsequences within S. If two subsequences match, we have found a pattern in the time series that is repeated. This pattern is defined as a motif.

Definition 4. Motif. A subsequence from T that is seen to repeat at least once throughout T is defined as a motif. The re-occurrence of the subsequence need not be exact for it to be considered as a motif.

The relationship between two subsequences $C_1$ and $C_2$ is assessed using a match threshold r. We use the most common distance measure (Euclidean distance) to examine the match between two subsequences $C_1$ and $C_2$, $ED(C_1, C_2)$. If $ED(C_1, C_2) \leq r$ the subsequences $C_1$ and $C_2$ are deemed to match and thus are saved as a motif.

The motifs prevalent in a time series are detected by the MTA through the evolution of a population of trackers.

Definition 5. Tracker. A tracker represents a signature for a motif sequence that is seen to repeat. It has within it a sequence of 1 to w symbols that are used to represent a dimensionally reduced equivalent of a subsequence. The subsequences generated from the time series are converted into a discrete symbol string using an intuitive technique described in Section 5.1. The trackers are then used as a fast and efficient tool to identify which of these symbol strings represent a recurring motif. The trackers also include a match count variable to indicate the level of stimulation received during the matching process.

## 5 The Motif Tracking Algorithm

The pseudo code for the MTA is detailed in Program 1. Each of the significant operations in the MTA is described in the subsequent sections. The parameters required in the MTA include the length of a symbol s, the match threshold r, and the alphabet size a.

### 5.1 Convert Time Series T to symbolic representation

In common with the majority of other techniques[20], in order for the MTA to more easily identify motifs it is necessary to pre-process the time series. This pre-processing comes in three stages, differencing, normalisation, and symbolisation.

Differencing. The approach used in the MTA is not to study the actual data values from T at each point in time but to look at the movement between each point in time. By taking the first order differential of the time series T, the MTA can identify patterns that occur in sections of the time series that have different amplitudes. Differencing the time series T eliminates discrepancies caused by the amplitude of the actual data points and hence aids comparability of the various subsequences.

Normalisation. The differenced time series T is normalised with a mean of 0 and a standard deviation of 1. Keogh highlights the fact that "it is well understood that it is meaningless to compare time series with different offsets and amplitudes"[20]. Normalising T in this way eliminates these issues and allows comparisons across T.

Symbolisation. We use the Keogh's SAX technique[20] to discretise the time series under consideration. SAX represents a powerful compression tool that uses a discrete, finite symbol set to generate a dimensionally reduced version of a time series consisting of symbol strings. The ability to compare time series subsequences as simple strings, combined with the dimensionally reduced set of data to investigate, contribute to a fast and effective search mechanism. This representation is simple and intuitive and research by Keogh et al. has shown it to rival more sophisticated reduction methods such as Fourier transforms and wavelets[20]. Here we describe Keogh's SAX approach, as it has been applied to the MTA.

In the first stage of the SAX method the user is required to specify the symbol alphabet size a used to represent the time series T. For example, if we use the English alphabet as our symbol set and a=3, the alphabet for the MTA would be [a,b,c].

Motifs in T consist of subsequences of lengths from 1 to n. A subsequence will be represented by a symbol string containing w symbols. By specifying the length of a symbol s, we can reduce the subsequence from size n to size w, where w=n/s.

This simplification is achieved by a Piecewise Aggregate Approximation (PAA)[9]. The n consecutive data points representing the motif are divided into w equal sized frames where w = n/s. The mean value of the data within each frame is calculated and represents the PAA of that frame. The motif now consists of a sequence of w averages corresponding to each frame. These averages are converted into the symbol alphabet as part of the second stage of the SAX.

During the second stage, the MTA calculates the break points it will apply to the averages to determine the symbol

---

Program 1. MTA Pseudo Code

```
Initiate MTA (s, r, a)
Convert Time series T to symbolic representation
Generate Symbol Matrix S
Initialise Tracker population to size a
While ( Tracker population > 0 )
{
    Generate symbol stage matrix from S
    Match trackers to symbol stage matrix
    Eliminate unmatched trackers
    Examine T to confirm genuine motif status
    Eliminate unsuccessful trackers
    Identify and store motifs found
    Proliferate matched trackers
    Mutate matched trackers
}
Memory motif streamlining
```

applicable for that average. Given an alphabet size a the MTA requires a-1 breakpoints to classify the averages correctly. Since T has been normalised we know T will follow a Gaussian distribution[20]. Given a symbol alphabet of size a we can use the Gaussian distribution to identify a equal sized areas under the curve. The boundary points for each equal sized area will then map directly to the break points applied to each frame average to establish which symbol corresponds to that average.

For example given a = 3, we split the area under the N(0,1) Gaussian distribution curve into three equally sized parts. One third of the distribution lies below $z=-0.43$, one third lies between $z=-0.43$ and $z=0.43$ and the final third exists above 0.43. We use the two breakpoints of -0.43 and 0.43 to classify our PAA values for each frame into symbols. PAA values less than -0.43 would correspond to the symbol a, those between -0.43 and 0.43 would represent symbol b, and those above 0.43 c. In this way each symbol has an equi-probable chance of occurring.

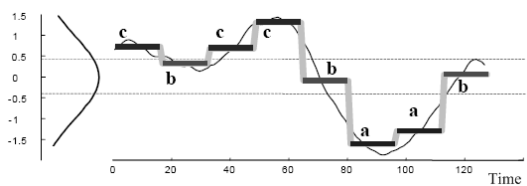

Figure 1: Generation of the SAX representation of a time series using PAA[18]. The X axis is time, the Y axis is the differenced, normalised time series value

This process is illustrated in Fig. 1, as taken from[18], where we have a subsequence of length 128, a = 3 and s = 16. Using the averages generated by the PAA, SAX represents the subsequence by the symbol string cbccbaab. The motif has been reduced in dimension from 128 data points to just 8 whilst still maintaining the characteristics of the original data.

The string of symbols representing a subsequence is defined as a word. The MTA examines the time series T using such words and not the original data points to speed up the search process. Symbol string comparisons can be performed efficiently to filter out bad motif candidates, ensuring the computationally expensive Euclidean distance calculation is only performed on those motif candidates that are potentially genuine.

## 5.2 Generating symbol matrix S

Thus far we have differenced and normalised the time series T and established the alphabet of symbols to be used to represent T. The next step is to identify all the actual individual symbols that occur across T.

The MTA begins by considering subsequences of length s. A sliding window of length s is used on the normalised time series to establish the PAA value for that subsequence. A symbol is allocated to that subsequence and entered into the symbol matrix S as a word. Given there will be (m-s+1) sliding windows across T the matrix S will comprise a list of (m-s+1) words containing just one symbol.

The MTA uses the SAX approach to generate the matrix of words, but this represents only the first stage of the MTA's operation. The SAX technique is only used to pre-process the data set. Having generated this symbol matrix the novelty of the MTA comes from the way in which it takes the information from the symbol matrix and intuitively presents it to the tracker population for matching. The way in which the tracker population evolves based on the results of the matching process is also key to the MTA's success.

The matrix S provides a vital resource to the MTA as it is used to build the data that is presented to the tracker population. Each generation, a selection of words from S, corresponding to the length of the motif under investigation, are extracted in an intuitive manner and presented to the tracker population during the matching process. Before describing this matching process we now expand on the concept of a tracker.

## 5.3 Initialise tracker population to size a

The trackers are the primary tool used to identify motif candidates in the time series. A tracker comprises a sequence of 1 to w symbols. The symbol string contained within the tracker represents a sequence of symbols that are seen to repeat throughout T. The level of stimulation of a tracker is indicated by the match count variable which is set to zero on initialisation.

Tracker initialisation and evolution is deterministic in nature to avoid proliferation of ineffective motif candidates. Given the user defined alphabet size a, the initial tracker population is constructed of size a to contain one of each of the viable alphabet symbols. Each tracker is unique, to avoid unnecessary duplication of solution candidates and wasted search time.

Trackers are created of a length of one symbol, but must evolve each generation to match motifs represented by longer symbol strings. The trackers are matched to motifs via the words extracted from the symbol matrix. Trackers that match a word are stimulated by incrementing their match counts by one; trackers that attain a match count $\geq$ 2 indicate repeated words from T and become candidates for proliferation. Given a motif and a tracker that matches part of that motif, proliferation enables the tracker to extend its length by one symbol each generation until it their lengths are equal, in order for it to attempt a match to the whole motif.

## 5.4 Generate symbol stage matrix from S

Initially the tracker population contains trackers that are one symbol long. The symbol matrix S contains a time ordered list of all symbols present in the time series T. We could now match the trackers to the words in S to identify potential motif candidates.

However, S contains a considerable amount of redundant information. Keogh defines such redundant data as trivial matches. Given a time series T, containing a subsequence C beginning at p and a matching subsequence M beginning at q, M is considered a trivial match to C if either p = q or there does not exist a subsequence M' beginning at q' such that ED(C,M')>r, and either q<q'<p or p<q'<q[18].

Trivial matches are sequences that are located in consecutive locations within S and so match each other. Such trivial matches should be eliminated from the search to avoid processing of unnecessary information. Given the similarity of neighbouring sliding windows we only wish to examine data from those windows that generate non trivial sequences.

A stage matrix is created from S each generation and its primary purpose is to eliminate such trivial matches. The MTA considers each symbol in S and only transfers it to the stage matrix if that symbol differs from the previous symbol entered.

The risk of trivial match elimination is that if the symbol set size a used to represent the time series is too small, we will get excessive elimination of consecutive sequences that are not trivial. Non trivially matching sequences may be represented by the same symbol, given the limited set available, leading to inappropriate eliminations taking place. To prevent this occurring the maximum number of consecutive trivial match eliminations must be less than s. In this way a subsequence can eliminate as trivial all subsequences generated from sliding windows that start in locations contained within that subsequence but no others. This prevents excessive symbol amalgamation and the loss of data.

For example, if the symbol size was s = 3 and the symbol matrix contained the words [c,c,c,c,d] full trivial match elimination would generate the stage matrix [c,d]. Restricting the trivial match elimination to s = 3, only the second and third words are removed as trivial to leave [c,c,d]. This is important as the first and fourth words may indicate sequences that are a true match for each other and correspond to a motif.

This reduction does not mean a loss in data quality if done carefully. The start location of each subsequence is stored within the word corresponding to the subsequence so that the range of trivial matches can be recalled. For example, if two consecutive words within the stage matrix were (a, 100) and (c, 106), the first entry corresponding to the subsequence symbol string and the second to the location, the MTA could identify the subsequences starting from 101 to 105 as all having the symbol a. Trivial match elimination in this case ensures that the five subsequences from points 100 to 105 are not compared against each other as they reflect potential trivial matches.

The stage matrix is then presented to the tracker population for matching.

### 5.5 Match trackers to the symbol stage matrix

During an iteration each tracker in the tracker population is taken in turn and compared to each word within the stage matrix. Matching is assessed by extracting and comparing the symbol strings from the tracker and the word. We define a match to occur if the comparison function returns a value of 0, indicating a perfect match exists between those symbol strings.

Each time the tracker is found to match a word in the stage matrix the tracker is stimulated by incrementing its match counter by one. Trackers with a match count $\geq 2$ indicate words that have reoccurred throughout the time series, allowing the MTA to narrow the search to these potential motif candidates.

### 5.6 Eliminate unmatched trackers

During this step the MTA eliminates all trackers that have a match count of zero or one as they do not satisfy the definition of a motif per Section 4. They correspond to symbol combinations that either only occurred once in the time series or that did not occur at all. Match counts for each tracker are then reset for the next matching stage.

For example, suppose a = 3, so the initial tracker population is [a,b,c], and the stage matrix consisted of the words [a,a,b,b,a,b]. After symbol matching the only trackers to be stimulated will be a and b. The trackers identify the sequences in the time series that are seen to repeat. c represented a potentially viable motif component but it was not observed in the symbol matrix. The tracker corresponding to c receives no stimulation and will be eliminated as redundant. Knowledge of possible motif candidates from T is therefore maintained and carried forward by the tracker population. Eliminating non repeating trackers ensures the MTA is only focused on investigating viable motif candidates.

### 5.7 Examine T to confirm genuine motif status

For each surviving tracker the MTA then scans through the stage matrix looking for two subsequences X and Y whose symbol strings correspond to the current tracker. However, even if X and Y have the same symbol strings they may not represent a true match when looking at the time series data underlying those subsequences.

In order to confirm whether the two subsequences X and Y correspond to genuine motifs in T we need to apply a distance measure on the first order differential of the time series associated with those subsequences and compare it to the match threshold r. The MTA uses Euclidean distance to measure the relationship between two motif candidates ED(X,Y).

To ensure the subsequences are a close match across the whole of their length, and represent genuine motifs, the subsequences X and Y were separated into w subsets, $x_{1..w}$ and $y_{1..w}$. Here w = the number of symbols in the word, so each subset corresponds to a symbol from the word representing that subsequence.

For simplicity we define these subsets as the symbol subsets of a subsequence, each of which has the length equal to the symbol size s. The Euclidean distance calculation is then applied on each of these symbol subsets in turn. As soon as the $ED(x_{1..w}, y_{1..w})$ returns a value $> r$ the subsequences X and Y are rejected as a potential motif candidate. This creates a strict match criteria.

If the Euclidean distance of all symbol subsets relating to X and Y are $\leq r$ a motif has been found. A memory motif is created to store the symbol string associated with X and Y, in addition the start locations of X and Y from T are also saved. The match count of the current tracker is then incremented by one. The MTA then continues to search for other words from the stage matrix which have the same symbol string as that tracker to identify further occurrences of that motif.

Having identified all occurrences associated with the currently found motif, the MTA recommences its search of the stage matrix from subsequence X to identify the next subsequence Z to match the current tracker. Once found the MTA follows a similar search mechanism as described above to identify all possible matches to Z.

Setting a correct value for r is essential to the operation of the MTA. A static r would be unsuitable for the MTA because during each generation the trackers and words from the symbol matrix that are matched are lengthened by one symbol. It is inappropriate to compare matches over longer sequences with the a static Euclidean distance threshold. The value of r has to dynamically adjust to be proportional to the length of the motif under consideration to avoid the premature dismissal of longer, still valid motifs.

To resolve this issue a distance threshold D is established per time point, based on a percentage of the standard deviation of the first order differential of T. Given the user defined symbol length s, the MTA calculates a match threshold r = D.s. This ensures a dynamic and flexible match threshold is established that is applicable across alternative time series.

### 5.8 Eliminate unsuccessful trackers

During step 2 of the tracker matching process, trackers that identified genuine motifs in T had their stimulation factors incremented. The MTA once again eliminates all unstimulated trackers to allow the MTA to narrow down motif solutions quickly and avoid processing unnecessary search candidates.

### 5.9 Identify and store motifs found

The motifs identified during stage 5.7 are stored in the memory pool for review. Comparisons are made to ensure duplications are removed. This memory pool represents the compressed representation of the time series, containing all the re-occurring patterns present.

### 5.10 Proliferate matched trackers

After all elimination steps the tracker population consists of symbols with underlying time series subsequences that are seen to repeat throughout the time series. They therefore correspond to parts of, or the whole of, the motifs we wish to find. During early iterations of the MTA the length of the trackers will be less than the complete motifs they have to find, since trackers are initialised to a length of one symbol. In order for the trackers to lengthen and therefore capture more of the motif the MTA lengthens the surviving trackers in a controlled manner through directed proliferation and mutation.

At the end of the first generation the surviving trackers, each consisting of a single symbol, represent all the symbols that are applicable to the motifs in T. The full motifs in T can only consist of combinations of these symbols. As such, at the end of generation one, the surviving tracker population is stored as the mutation template for the MTA. If we are to undertake proliferation and mutation of the trackers it makes sense that all mutations should only involve symbols from the mutation template and not the full symbol alphabet, as any other mutations would lead to unsuccessful motif candidates.

The MTA takes each surviving tracker in turn and proliferates it to generate a number of clones equal to the size of the mutation template. These clones adopt the same symbol string as their parent.

### 5.11 Mutate matched trackers

The clones generated from each parent are taken in turn and extended by adding a symbol taken consecutively from the mutation template. Once the clones are mutated they are added back into the tracker pool. This results in a unique population of trackers with maximal coverage of all potential motif solutions and no duplication. This deterministic proliferation and mutation of the tracker pool [a,c,d] is illustrated in Fig. 2.

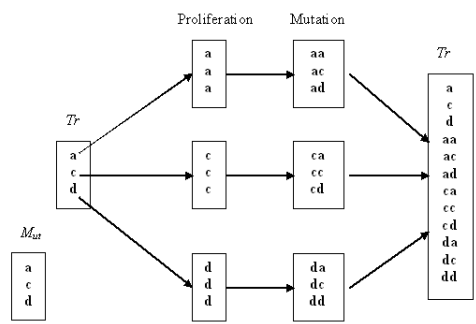

Figure 2: A visual illustration of the proliferation and mutation process applied to the tracker population [a, c, d]. $M_{ut}$ = the mutation template and Tr = the tracker population.

Post proliferation and mutation, the tracker pool is fed back into the MTA ready for the next generation. The surviving trackers now include symbol strings of length two, as well as one. In the second generation a new stage symbol matrix must be formulated to present to these evolved trackers, consisting of motif candidates of two symbols. In this way the MTA builds up the representation of a motif one symbol at a time each generation to eventually map to the full motif.

Given the symbol length s we can generate a word consisting of two consecutive symbols by taking the symbol from matrix S at position i and adding to it the symbol from position i+s. Repeating this across S the MTA obtains a new stage matrix in generation 2, each entry of which contains a word of two symbols, covering a subsequence of length 2s.

Selection from the symbol matrix S into the new stage matrix is performed with trivial match elimination as described in Section 5.4 to reduce the set of alternative solution candidates. Given the symbol strings are now longer and more specific trivial match elimination will remove fewer subsequences. However, invalid subsequences from amongst these alternatives are being quickly dismissed via the symbol string matching mechanism as trackers evolve to map to the longer motif candidates. The MTA is therefore elegantly narrowing down its search path as it hunts for genuine motifs.

The MTA continues to prepare and present new stage matrix data to the evolving tracker population generation after generation until all trackers are finally eliminated as non matching. Once the tracker population size falls to zero the MTA stops. Any further extension to the current tracker population will not improve their fit to any of the underlying motifs in T.

## 5.12 Memory motif streamlining

The final stage of the MTA analyses the memory motif population and eliminates those motifs that are found to be completely encapsulated within other motifs. In addition similar motifs are combined if upon investigation they are found to separately represent parts of the same repeating pattern. The MTA streamlines the memory pool to generate a list of individual motifs for T, with no duplication. Each memory motif consists of the symbol string epitomising the general characteristics of the motif, the motif length and the location of each occurrence of the motif.

# 6 Results

Having introduced the MTA we now move provide some initial results which examine the ability of the MTA to identify motifs.

## 6.1 Finding motifs embedded in a random walk

To ensure the MTA works as intended two randomly generated motifs A and B were embedded into a random walk data set of length 400. Motif A consists of a sequence of length 40 embedded at time points 47 and 160. Motif B consists of a sequence of length 40 embedded in T at time points 100 and 230. A and B were chosen such that they are realistic enough for validation. Fig. 3 illustrates the time series T generated and the embedded motifs A and B.

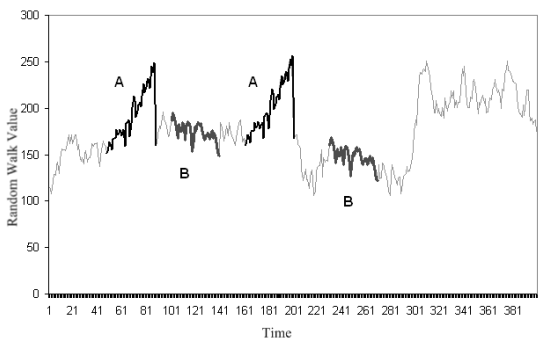

Figure 3: Finding the known motifs A and B embedded in a random walk time series

In this simple test scenario $a = 6$, $s = 10$, and $r = 0.5$ per time point. These parameters were established as suitable after numerous runs of the MTA. Sensitivity analysis on these parameter settings is investigated in Section 6.3.

The MTA analysed T in 14 seconds and found evidence of two motifs. The first motif, represented by the word dcdd, was found located at points 47 and 160 and was of length 41. This motif exceeded the 40 day length that was originally embedded because the data from points 87 and 200, found immediately after the embedded motif, was also very similar. The length of the motif found was therefore appropriately extended by one day to include this matching data. The second motif, dccc, was found located at days 100 and 230 and was of length 40. Both embedded motifs were found successfully while no other motifs from the non embedded sections of the random walk were identified. This simple test indicates the MTA is capable of identifying motifs present in a time series.

## 6.2 Finding unknown motifs in real world data

The previous example shows the MTA can uncover motifs in a small random walk data set, but the aim of the MTA is to find motifs in larger real world time series. The dataset selected for investigation in this scenario is the steamgen data set. This data was generated using fuzzy models[25] applied to the model of a steam generator at the Abbott Power Plant in Champaign[26] and is available from http://homes.esat.kuleuven.be/~tokka/daisydata.html. The data output from the models consists of four measured variables: drum pressure, excess oxygen levels, water levels and steam flow. The steamgen data set consists of every tenth observation taken from the steam flow output information, starting with the first observation. This specific selection criterion was used by Keogh and has been followed here for the purposes of comparison. The steamgen data set contains 960 items with significant amplitude variation, a subset of 400 of which is illustrated in Fig. 4.

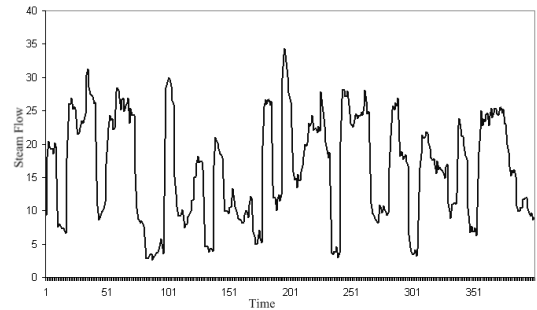

Figure 4: A subset of 400 data items from the steamgen data set, provided as a visual context for the motif detection problem. The X axis refers to Time, whilst the Y axis refers to steam flow.

From an initial scan by eye of the data it is unclear whether any significant motifs exist, representing an ideal challenge for the MTA. The three parameters s, a and r need to be determined for the MTA. We will examine the sensitivity of the MTA to changes in these parameters, but for now we set $s = 10$, $a = 6$ and $r = 0.5$.

Using these parameters, the MTA identifies 104 motifs of varying lengths between ten and 60. Some of the motifs of length ten are seen to repeat up to 15 times throughout the data, others of length 20 are noted to repeat up to four times. However one significant motif of length 60, seen to occur twice in the data, dominates the motif pool. The subsequences comprising this motif, starting at location 75 and 883, are remarkable in their similarity as can be seen in Fig. 5.

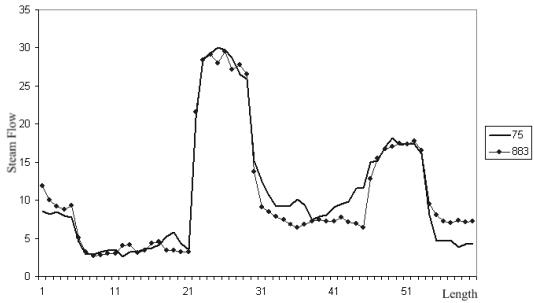

Figure 5: The plot of a motif found in the steamgen data by the MTA. It consists of the subsequences starting at locations 75 and 883, both of length 60. The X axis refers to the motif length, whilst the Y axis refers to steam flow.

This result implies that the MTA can identify unknown motifs in this real world data. One could ask however how do we know we have captured all of this motif? In order to provide some grounding for the MTA, we compared the MTA result to that of the probabilistic motif search algorithm used by Keogh et al.[18]. Keogh was kind enough to provide a teaching version of his algorithm which we applied to the steamgen data, using parameters established by Keogh for this data set. Running the probabilistic algorithm, specifying a search for motifs of length 80, results in an obvious motif being found. The motif discovered consisted of sequences starting at points 66 and 874, covering a length of 80. This motif is illustrated in Fig. 6.

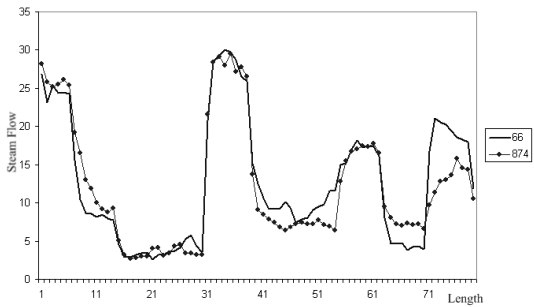

Figure 6: The plot of a motif found in the steamgen data by Keoghs probabilistic algorithm. It consists of the subsequences starting at locations 66 and 874, both of length 80. The X axis refers to the motif length, whilst the Y axis refers to steam flow.

Comparing Figures 5 and 6 it appears that the MTA has only detected a subset of the motif found by the probabilistic algorithm, missing off the first and last ten data points of the longer motif. The Euclidean distances for the first and last ten days of the 80 day motif are 5.48 and 11.17 respectively. We know the MTA applies a Euclidean distance threshold across subsets of the motif corresponding to each symbol, to ensure any variation in the sequences under comparison is evenly distributed.

Remembering $s = 10$, and $r = 0.5$ per unit, we have a match threshold for each ten unit period of 5.0. The Euclidean distances of 5.48 and 11.17 both exceeded this match threshold. Therefore the MTA considered these regions during its investigation but dismissed them as not matching. Comparing Figures 5 and 6 this is clearly evident. The MTA is more stringent in its definition of a match and has identified the same motif as the probabilistic algorithm but only that part that represents the closer match.

Establishing that the MTA can find unknown motifs in a real world data set, and having those motifs confirmed through comparison to an alternative motif detection algorithm provides a good grounding for the MTA. A further important consideration however is to test the sensitivity of the MTA to changes in the parameters s, r and a.

## 6.3 The MTA and parameter sensitivity

Using the steamgen dataset the MTA was run multiple times with variations of the parameters s, r and a to examine the impact on the final motif population, and the ability of the MTA to retain knowledge of the motif shown in Fig. 5. We define this motif $M_1$. Values of $s = 10$, $a = 6$ and $r = 0.5$ per unit were set as base line defaults for each experiment. After each experiment the following information was recorded in Table 1.

C1 The number of motifs.

C2 The number of repetitions of all motifs.

C3 The average motif length.

C4 The standard deviation of the motif length.

C5 The percentage of T covered by all motifs.

C6 The average euclidean distance of all motifs.

C7 The MTA execution time in milliseconds.

C8 The length of the version of motif $M_1$ found.

Given this information we can determine a measure of the quality of the motifs, MQ, found with the following formula.

$$MQ = (C2 * C3)/C6 \qquad (1)$$

MQ provides a measure of the size and fitness of the motif population that is found. A large MQ value is beneficial as it indicates the MTA finds more occurrences of longer motifs that are a closer match to each other.

In addition we can attain a measure of the efficiency of the algorithm, ME, in terms of execution time, from the formula

$$ME = (C5/C7) \qquad (2)$$

ME provides a measure of the performance of the MTA, it assesses how long the algorithm takes to find additional motifs that are of value. ME highlights the trade off that exists between the size of the motif population found and the execution time of the MTA, so we can ensure the solution is reached efficiently. A rise in ME implies it takes less time to find the same motif coverage of T. The results of the parameter testing are listed in Table 1.

Table 1 Sensitivity analysis of the MTA to changes in parameters s, r and a.

| Sensitivity of changes to symbol size 's' | | | | | | | | | | | |
|---|---|---|---|---|---|---|---|---|---|---|---|
| s | r | a | C1 | C2 | C3 | C4 | C5 | C6 | C7 | C8 | MQ | ME |
| 5 | 0.5 | 6 | 254 | 2,318 | 5.5 | 1.8 | 96.5 | 2.0 | 1,113,752 | 15 | 6,519 | 0.1 |
| 10 | 0.5 | 6 | 104 | 428 | 12.2 | 5.5 | 92.7 | 4.3 | 127,173 | 60 | 1,203 | 0.7 |
| 15 | 0.5 | 6 | 75 | 285 | 18.7 | 8.7 | 92.6 | 6.4 | 68,769 | 70 | 830 | 1.3 |
| 20 | 0.5 | 6 | 50 | 140 | 27.2 | 13.1 | 90.1 | 8.8 | 33,649 | 68 | 431 | 2.7 |

| Sensitivity of changes to match threshold 'r' | | | | | | | | | | | |
|---|---|---|---|---|---|---|---|---|---|---|---|
| s | r | a | C1 | C2 | C3 | C4 | C5 | C6 | C7 | C8 | MQ | ME |
| 10 | 0.3 | 6 | 23 | 69 | 10.3 | 1.7 | 36.8 | 2.7 | 20,710 | 10 | 263 | 1.8 |
| 10 | 0.4 | 6 | 61 | 225 | 10.5 | 2.1 | 77.1 | 3.4 | 46,977 | 20 | 686 | 1.6 |
| 10 | 0.5 | 6 | 104 | 428 | 12.2 | 5.5 | 92.7 | 4.3 | 127,173 | 60 | 1,203 | 0.7 |
| 10 | 0.6 | 6 | 150 | 735 | 13.2 | 6.1 | 97.6 | 5.1 | 262,868 | 63 | 1,893 | 0.4 |
| 10 | 0.7 | 6 | 215 | 1,076 | 14.5 | 7.6 | 99.6 | 6.0 | 470,407 | 63 | 2,601 | 0.2 |
| 10 | 0.8 | 6 | 257 | 1,395 | 15.3 | 8.2 | 100.0 | 6.9 | 720,756 | 63 | 3,091 | 0.1 |

| Sensitivity of changes to alphabet size 'a' | | | | | | | | | | | |
|---|---|---|---|---|---|---|---|---|---|---|---|
| s | r | a | C1 | C2 | C3 | C4 | C5 | C6 | C7 | C8 | MQ | ME |
| 10 | 0.5 | 4 | 106 | 438 | 11.8 | 4.4 | 93.6 | 4.3 | 97,691 | 40 | 1,201 | 1.0 |
| 10 | 0.5 | 6 | 104 | 428 | 12.2 | 5.5 | 92.7 | 4.3 | 127,173 | 60 | 1,203 | 0.7 |
| 10 | 0.5 | 8 | 124 | 609 | 12.1 | 4.8 | 96.1 | 4.3 | 188,101 | 40 | 1,699 | 0.5 |
| 10 | 0.5 | 10 | 119 | 615 | 11.7 | 4.8 | 94.8 | 4.3 | 211,294 | 50 | 1,692 | 0.4 |

### 6.3.1 Changes to symbol size s

Given the default values of a = 6, r = 0.5, s was examined for the values 5, 10, 15, and 20.

With low s values the MTA finds a large number of short motifs, which occur very frequently, all of which have a good general fit. However the MTA is unable to find long motifs. The average motif length is 5.5 days when s=5, and only 15 days of motif $M_1$ could be found. With low s values the MTA execution time is significantly longer, reaching a maximum of 1,113 seconds (C7).

Increasing s raised the average Euclidean distance of the motif solutions (C6). This is to be expected given the match threshold (r.s) is proportional to s. The longer the symbol size, the higher is the match threshold, leading to solutions with a higher Euclidean distance being found. Effectively the MTA is more lenient regarding the acceptance of motif candidates.

Fewer motifs are found as s rises, but those found constitute longer motifs. The average motif length rises from 5.5 to 27 days (C3), with a standard deviation rising from 1.8 to 13 days (C4). This indicates much more variety in the length of motifs found. The loss in the number of motifs detected occurs because some shorter motifs are combined into larger motifs, whilst other are simply missed by the MTA. The loss of motifs is shown by the slight fall in the motif coverage percentage (C5) as s rose.

In contrast, the execution time of the MTA significantly improves as s increases (C7), leading to an improvement in the efficiency measure ME.

Once s rises above ten the full length of 60 days for the motif $M_1$ is found, further increases to s offer little improvement or loss in the detection of this motif.

In summary, low s values result in excessive numbers of short motifs being found and long motifs being omitted. The short motifs are then either combined into longer motifs or are lost as s rises. Excessively high s values improve execution time but leaves the MTA unable to find some of the shorter motifs. A rise in s over ten offers no real benefit in the detection of the specific motif $M_1$.

### 6.3.2 Changes to the match threshold r

Given the default values for a = 6, s = 10, r was examined for values 0.3, 0.4, 0.5, 0.6, 0.7 and 0.8.

Low r values (0.3) provide an excessively strict match condition, resulting in only 23 motifs being found. These 23 motifs represent high quality matches but they cover only 36.8% (C5) of T. A data point identified as belonging to a motif is flagged. The percentage coverage is then calculated as the number of data points flagged over the total time series size. Given so few motifs are found the execution time for the MTA is minimal, taking only 20 seconds to complete.

High r values (0.8) result in an over-generous matching condition which extends the MTA's execution time to over 720 seconds, reducing the efficiency measure ME to 0.14. A large number of less well fitting motifs are now found, which cover 100% of T, indicating the match condition is too lenient.

A rise in r has less of an impact on the average motif length as that seen with s. Average motif lengths only rose from 10 days to 15 days as r rose from 0.3 to 0.8. Once r reaches 0.5 the MTA is able to identify all 60 days of motif $M_1$ and any further increases in r do not improve on this.

Once r rises to 0.7 a new motif $M_2$ is found of length 63, consisting of subsequences from points 118 and 774. This new motif is illustrated in Fig. 7. One can immediately see the subsequences in $M_2$ are a poorer fit to each other than those of $M_1$, as seen in Fig. 5. This is confirmed by the fact that $M_2$ has a 39% higher Euclidean distance than $M_1$.

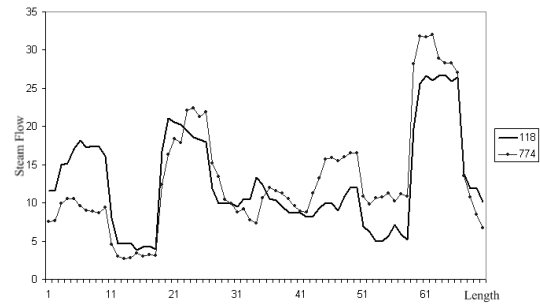

Figure 7: Motif $M_2$ found in the steamgen data set, consisting of subsequences of length 80 starting at locations 118 and 774. The X axis refers to the motif length, whilst the Y axis refers to steam flow.

From this analysis we conclude that increasing r causes a significant deterioration in the efficiency of the MTA as the overgenerous match threshold entails comparisons be made across a larger number of poorer motif candidates. The result is a poorly fitting motif pool. However, excessively low r values exclude the detection of most motifs. An r value in between these extremes is desirable, such as r = 0.5, where motifs, such as $M_1$, are detected without loss and any further increases offer no improvement to the solution.

### 6.3.3 Changes to the alphabet size a

Given the default values for s = 10, r = 0.5, a was examined for values 4, 6, 8, and 10.

From Table 1 it can be seen that changing a has little impact on the average Euclidean distance (C6), the average motif length (C3), the standard deviation of the motif length (C4), or the percentage of T covered by the motif population (C5).

Increasing a from 4 to 10 triggers an increase in the number of motifs detected from 106 to 119, and the number of motif repeats from 438 to 615, this in turn causes a 116% rise in execution time, leading to a fall in the efficiency measure ME to 0.45.

More motifs are found as a rises because of a decline in the number of words eliminated during trivial match elimination. With small a values T is represented by a smaller alphabet set. There is less variety in the symbol alterna- tives used to represent the motif subsequences, leading to greater trivial match elimination. But as a rises the symbol set increases, resulting in fewer potential trivial matches in S. Looking at the size of the symbol stage matrix in gen- eration one after trivial match elimination we see when s = 4, the stage matrix is reduced from 960 data items to 176 through trivial match elimination, alternatively if s = 10 the data set is reduced to only 299. The larger degree of trivial match elimination causes the MTA to miss some motifs.

Unlike changes in s or r, changes to a have less of an impact on the detection of motif $M_1$. The MTA is always able to identify a length for $M_1$ of at least 40, regardless of a. Looking at the graph of $M_1$ in Fig. 5 one can see the first 40 time periods indicate the closest match for the subsequences comprising $M_1$. Changing s never effects the detection of this closely fitting region, it only changes the detection of the less well fitting region in the rest of the motif.

From all this information one can hypothesize that changing a only has a limited impact on the quality and quantity of the motif population, however further analysis of a still needs to be performed to ensure the alphabet size chosen does not significantly influence the MTA's accuracy.

The analysis of parameter sensitivity shows that while the MTA remains relatively insensitive to changes in a, it is sensitive to changes in both r and s. This is understandable given s and r have a direct effect on the matching criteria applied to subsequences when testing to see if they form a motif. As both parameters influence the match threshold, a rise or fall in either would result in a respective relaxing or tightening of the matching criteria. The choice of s and r will therefore be influenced by the data set under examination. However the advantage of the MTA, when compared to other algorithms, is that this dependency is just limited to setting the parameters s and r. In future work we propose to normalise the subsequences used in the Euclidean distance calculations so the sensitivity of r to different data sets could be reduced. This leaves the user only having to select an appropriate value for s, making the algorithm flexible across different problem domains. Sensitivity analysis of the MTA parameters in other problem domains has also been investigated[10].

It would have been beneficial to compare the statistics generated from the MTA to the equivalent of Keogh's probabilistic model. However this was difficult given the output format generated by Keogh's algorithm. The probabilistic algorithm generates a collision matrix of all motifs in a significantly quicker time frame than the MTA. This provides a nice visual representation of the motifs found but it is difficult to compile a comprehensive list of the exact number of motifs, their affinity and their frequency. In addition the motif length has to be specified in advance by the user making the probabilistic search very specific. However one can gain comfort from the fact that the significant motifs found by the MTA were also found by Keoghs algorithm and vice versa.

### 6.4 Motifs in oil data

The previous sections indicate that the MTA is able to identify unknown motifs in small data sets and real world data.

Of particular interest to us is the application of the MTA to a financial time series. Stock market data is a popular selection for many pattern matching algorithms, given the financial rewards of successfully forecasting future price movements based on historical information. One of the most important stock market commodities is that of oil, given its influence on the rest of the market. For this reason we have chosen to apply the MTA to the daily West Texas Intermediate (WTI) crude oil price, a popular indicator of general oil prices. This data spans the period covering January 1986 to January 1990 and contains 1020 items, as plotted in Fig. 8 below.

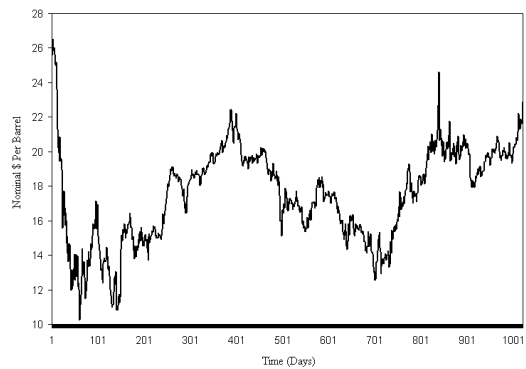

Figure 8: The daily WTI crude oil spot price Jan '86 to Jan '90 measured in dollars per barrel

From an initial scan by eye of Fig. 8, one could hypothesize that the oil data has characteristics similar to that of a random walk. As a stock market commodity this is unsurprising as 'it is well known that random walk data can perfectly model stock market data in terms of all statistical properties, including variance, autocorrelation, stationarity'[27]. One could argue therefore that it would be difficult to extract any meaningful motifs from such data.

The MTA was initialised using the parameters a = 8, s = 5 and r = 0.08. Section 6.3 indicated the MTA was relatively insensitive to a so a high alphabet size of 8 should avoid the issue of excessive trivial match elimination and provide a good representation for the time series. The data was evaluated in symbols of length 5 to provide a detailed examination of the data and ensure the subsequences comprising the motifs were of a good match for each other. The match threshold of 0.08 corresponds to 15% of the standard deviation of the first order differential of the oil data time series. The determination of 15% was carried forward from the sensitivity analysis performed in Section 6.3 which resulted in realistic results for the steamgen data set and this oil data set.

Using these parameters the MTA was able to identify 250 motifs, three lasting 20 days, seven lasting 15 days, 59 lasting ten days, with the remainder lasting five days. The ten day motifs were seen to repeat up to six times throughout the time series, whilst some five day motifs were seen to occur 25 times across the data set. The motifs found covered 83.42 percent of the whole data set and it took the MTA 683 seconds to identify them.

Two examples of the motifs found by the MTA are illustrated in Figures 9 and 10.

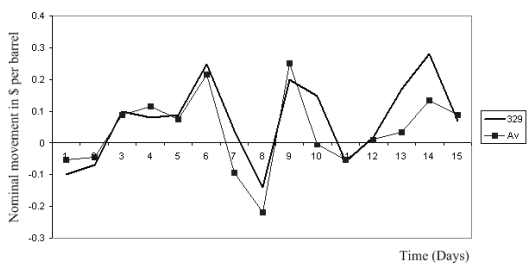

Figure 9: Motif $M_a$ of length 15, consisting of three subsequences found in the WTI oil dataset

The first motif, represented by the word eee, was found to be 15 days long and re-occurred three times in the time series, starting at days 329, 349 and 436. The nominal price change in dollars per barrel for the subsequence at day 329 is plotted against the average of the other two occurrences in Fig. 9. An immediate similarity is evident between these subsequences. This is confirmed by Euclidean distances between the subsequence 329 and those of 349 and 436 that both lie below 0.5 representing a close match.

The second motif, represented by the word ed, was found to be ten days long and re-occurred six times throughout the data, at days 214, 225, 359, 591, 720 and 919. Fig. 10 plots the subsequence starting at day 214 against the average of the other five occurrences. The Euclidean distance of each subsequence compared to 214 range from 0.34 to 0.46 indicating a close match once again.

From this simple analysis it is clear that motifs do exist in the WTI oil data set, and these can be found by the MTA. Having found these repeating patterns there is the potential to use the motif population as an additional resource to support other forecasting tools to predict future oil price movements.

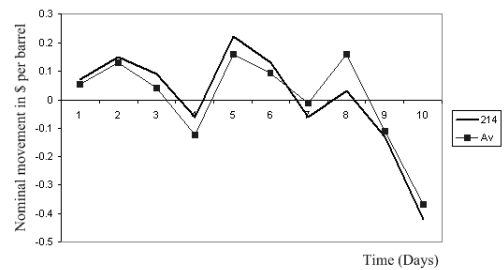

Figure 10: Motif $M_b$ of length 10 days, consisting of six subsequences found in the WTI oil dataset

## 6.5 The MTA and potential applications

The MTA represents a novel, abstract algorithm to identify unknowns motifs in a time series dataset in an intuitive and efficient manner. The population of motifs generated by the MTA is a potentially very useful resource that other algorithms could easily take advantage of.

Clustering and wavelet algorithms are usually seeded with random data upon initialisation. An alternative approach would be to seed these algorithms with the motif population generated by the MTA. The motifs represent known patterns that re-occur in the data set therefore giving the algorithms a head start in their analysis.

The motif generation process also represents a unique compression mechanism. The original time series is compressed to a reduced set representing the recurrent patterns in the data. This reduced set may be sufficient to provide a simple visual summary of the full data set.

Technical analysts working in the stock market use known and accepted patterns when analysing stock market performance. Such patterns include the 'head and shoulders' and 'cup and handle' patterns. The MTA could be applied to such stock market data and the motifs generated compared to these well established patterns. This analysis may be able to highlight other patterns that may also be of interest alongside those that are generally accepted.

A key potential application of the MTA would be to act as a support tool for forecasting. Executing the MTA on historical data would generate a population of motifs. When live data is received this information could represent a partial motif. Using principles from natural language one could compare the partial motif against the motif population and hypothesise the future direction or value for the partial motif. For example, suppose we took the oil data set described in Section 6.4 and assumed the last five days of live data translated to the word e. Review of all motifs containing the symbol e could provide an indication of what symbol is likely to reoccur next. The MTA found 77 motifs corresponding to the word e, these led on to 17 motifs consisting of ed, 15 of ee, 2 of ef and 1 of ec. One could therefore hypothesize that the next five days to come are likely to follow a pattern characterised by the symbols d or e.

## 7 Future work

A number of applications are listed in Section 6.5 and these could be incorporated into the MTA as part of our future work. The results presented in Section 6 are highly encouraging given the MTA is still in a relatively early stage of development. A number of modifications have been proposed and will be investigated as part of our future work. Having examined three separate data sets in this paper it is important to extend our analysis to evaluate the performance of the MTA across a wide range of diverse data sets. In this way we can ensure the technique used is successful and independence from the data set is maximised. Sensitivity analysis should be extended across these data sets. Performance measures need to be established for the MTA so that the algorithm can be benchmarked against other alternative approaches.

The incorporation of wild cards as an additional symbol in the available alphabet would be of interest as this mechanism may enable us to identify relationships between some of the motifs found. For example, consider we have a 10 day motif M found to occur on days 50 and 100, and a 20 day motif N was seen to occur on days 70 and 120. With a 10 day wildcard '*' one would immediately see there is a relationship between motifs M and N as we could effectively combine them into one motif 'M*N'.

In its current form the MTA makes no distinction of motif type during its search for motifs, it simply lists all those it finds. However users may only be interested in particular motif structures, for example financial analysts are likely to be interested in significantly fluctuating motifs and not stable, flat motifs. It would be of value to incorporate some concept of 'interestingness' from the user so that the MTA focuses its search on only this subset, reducing execution time and providing more meaningful solutions. Classifying motifs as 'interesting' has been applied with success by Keogh[28].

Alternative matching techniques other than Euclidean distance should also be investigated. Research by some authors has found that Euclidean distance performs well against other distance metrics on a wide variety of data sets[27], however it is important for alternative distance measures to be considered in the MTA to ensure its approach is valid and appropriate to the data.

## 8 Conclusions

The search for patterns or motifs in data represents a generic problem area that is of great interest to a huge variety of researchers. By extracting motifs that exist in data we gain some understanding as to the nature and characteristics of that data, so that we can benefit from that knowledge. The motifs provide an obvious mechanism to cluster, classify and summarise the data, in addition they can be used to predict future information, placing great value on these patterns.

Given the importance of finding motifs, considerable research has been performed in identifying known patterns in time series data. In contrast little research has been performed on looking for unknown motifs of an unspecified length that exist in time series. The MTA takes up this challenge using a novel immune inspired approach to evolve a population of trackers that seek out and match the motifs present in a time series. A key advantage of the MTA is that it uses a minimal number of parameters with minimal assumptions about the data examined or the underlying motifs when compared to alternatives approaches.

The MTA was evaluated using three data sets, one of which included oil price data, and in all cases the algorithm was able to identify the presence of a motif population. The unique search process of the MTA was aided through the utilisation of a simple and intuitive symbolic representation. Some of the motifs found were of considerable length or repeated with significant frequency to be of interest to a user of those time series. At present the MTA makes no distinction of 'interesting' motifs, it finds and presents all available motifs, but this could easily be addressed as part of our future work. In this work we highlight that fact that the MTA shows potential as a tool used for seeding other algorithms with its motif population to enhance their effectiveness, in addition the motif population could provide a valuable resource to aid in the forecasting of future information. The MTA is still at an early stage of development but the initial results presented in this paper are encouraging. We propose future work to improve and enhance this algorithm but even in its current form we believe that the MTA offers a valuable contribution to an area of research that at present has received surprisingly little attention.

## Acknowledgement

The authors would like to thank Eamonn Keogh from the Department of Computer Science and Engineering, University of California Riverside for providing a teaching version of his probabilistic motif determination algorithm, the steamgen dataset, and his advice and feedback.

## References

[1] M. Ghiassi, H. Saidane, and D. K. Zimbra. A Dynamic Artificial Neural Network Model for Forecasting Time Series Events. International Journal of Forecasting, 21:341–362, 2005.

[2] G. Zhang, D. E. Patuwo, and M. Y. Hu. Forecasting with Artificial Neural Networks: The State of the Art. International Journal of Forecasting, 14:35–62, 1998.

[3] C. Grosan, A. Abraham, S. Y. Han, and V. Ramos. Stock Market Prediction Using Multi Expression Programming, 2005. ALEA05, Workshop on Artificial Life and Evolutionary Algorithms at EPIA05.

[4] S. H. Chen. Genetic Algorithms and Genetic Programming in Computational Finance. Kluwer Academic Publishers: Dordrecht, 2002.

[5] I. Nunn and T. White. The Application of Antigenic Search Techniques to Time Series Forecasting. GECCO, pages 353–360, June 2005.

[6] J. H. Carter. The Immune System as a Model for Pattern Recognition and Classification. Journal of American Medical Informatics Association, pages 28–41, January 2000.

[7] L. N. de Castro and F. J. Von Zuben. Learning and Optimization Using the Clonal Selection Principle. IEEE Transactions on Evolutionary Computation, 6(3):239–251, 2002.

[8] T. Knight and J. Timmis. AINE: An Immunological Approach to Data Mining. In N. Cercone, T. Lin, and X. Wu, editors, IEEE International Conference on Data Mining, pages 297–304, San Jose, CA. USA, 2001.

[9] J. Lin, E. Keogh, S. Lonardi, and P. Patel. Finding Motifs in Time Series. In the 2nd Workshop on Temporal Data Mining, at the 8th ACM SIGKDD International Conference on Knowledge Discovery and Data Mining, July, 2002.

[10] W. O. Wilson, J. Feyereisl, and U. Aickelin. Detecting Motifs in System Call Sequences. In Proceedings of the 8th International Workshop on Information Security Applications (WISA 2007), to be published, 2007.

[11] E. B. Bell, S. M. Sparshott, and C. Bunce. CD4+ T-cell memory, CD45R Subsets and the Persistence of Antigen - a Unifying Concept. Immunology Today, 19:60–64, February, 1998.

[12] X. Guan and E. C. Uberbacher. A Fast Look Up Algorithm for Detecting Repetitive DNA Sequences. Pacific Symposium on Biocomputing, Hawaii IEEE Tran. Control Systems Tech., December 1996.

[13] G. Benson and M. S. Waterman. A Method for Fast Database Search for all K-Nucleotide Repeats. Nucleic Acids Res, 22(22):4828–4836, November 1994.

[14] I. Rigoutsos and A. Floratos. Combinatorial Pattern Discovery in Biological Sequences: TEIRESIAS Algorithm. Bioinformatics, 14 no. 1:55–67, 1998.

[15] E. Keogh and P. Smyth. A Probabilistic Approach to Fast Pattern Matching in Time Series Databases. In Proceedings of the Third International Conference on Knowledge Discovery and Data Mining, pages 20–24, 1997.

[16] C. Faloutsos, M. Ranganathan, and Y. Manolopoulos. Fast Subsequence Matching in Time Series Databases. In Proceedings of the SIGMOD Conference, pages 419–429, 1994.

[17] S. Singh. Pattern Modelling in Time Series Forecasting. Cybernetics and Systems - an International Journal, 31, issue 1, 2000.

[18] B. Chiu, E. Keogh, and S. Lonardi. Probabilistic Discovery of Time Series Motifs. SIGKDD, August, 2003.

[19] J. Lin, E. Keogh, and S. Lonardi. Visualizing and Discovering Non Trivial Patterns in Large Time Series Databases. Information Visualization, 4, issue 2:61–82, 2005.

[20] J. Lin, E. Keogh, S. Lonardi, and B. Chiu. A Symbolic Representation of Time Series, with Implications for Streaming Algorithms. In Workshop on Research Issues in Data Mining and Knowledge Discovery, San Diego, CA, pages 2–11, June, 2003.

[21] W. Wilson and S. Garrett. Modelling Immune Memory for Prediction and Computation. In 3rd International Conference in Artificial Immune Systems (ICARIS-2004), pages 386–399, Catania, Sicily, Italy, September 2004.

[22] A. S. Perelson and G. Weisbuch. Immunology for Physicists. Rev. Modern Phys., 69:1219–1267, 1997.

[23] D. Chowdhury. Immune Networks: An Example of Complex Adaptive Systems. In Artificial Immune Systems and their Applications, D. Dasgupta (ed), pages 89–104, 1999.

[24] A. Yates and R. Callard. Cell Death and the Maintenance of Immunological Memory. Discrete and Continuous Dynamical Systems, 1:43–59, 2001.

[25] J. J. Espinosa and J. Vandewalle. Predictive Control Using Fuzzy Models Applied to a Steam Generating Unit. Submitted for publication FLINS 98 third international workshop on fuzzy logic and intelligent technologies for nuclear science and industry, April 1998.

[26] G. Pellegrinetti and J. Benstman. Nonlinear Control Oriented Boiler Modeling, A Benchamrk Problem for Controller Design. IEEE Tran. Control Systems Tech., 4, No 1, January, 1996.

[27] E. Keogh and S. Kasetty. On the Need for Time Series Data Mining Benchmarks: A Survey and Empirical Demonstration. Data Mining and Knowledge Discovery, Volume 7, Number 4 / October, 2003:102 – 111, July 2002.

[28] E. Keogh, S. Lonardi, and B. Chui. Finding Suprising Patterns in a Time Series Database in Linear Time and Space. In the 8th ACM SIGKDD International Conference on Knowledge Discovery and Data Mining, pages 550–556, July 2002.